\def\year{2021}\relax
\documentclass[letterpaper]{article} 
\usepackage{aaai21}  
\usepackage{times}  
\usepackage{helvet} 
\usepackage{courier}  
\usepackage[hyphens]{url}  
\usepackage{graphicx} 
\usepackage{natbib}  
\usepackage{caption} 
\usepackage{url}
\usepackage{booktabs}
\usepackage{multirow, makecell}
\usepackage{tikz}
\usepackage{array}

\usepackage{amsmath, amsthm, amssymb}
\usepackage{cleveref}
\usepackage{longtable}
\usepackage{tikz}
\usepackage{booktabs}
\usepackage[all]{nowidow}
\usepackage{amssymb}
\newcommand{\PreserveBackslash}[1]{\let\temp=\\#1\let\\=\temp}
\newcolumntype{C}[1]{>{\PreserveBackslash\centering}p{#1}}
\newcolumntype{R}[1]{>{\PreserveBackslash\raggedleft}p{#1}}
\newcolumntype{L}[1]{>{\PreserveBackslash\raggedright}p{#1}}

\newcommand{\sL}{\mathcal{L}}
\newcommand{\bt}{{\bf t}}
\newcommand{\transformation}[1]{\texttt{#1}}
\newcommand\sM{\ensuremath{\mathcal{M}}}
\title{BERT \& Family Eat Word Salad: Experiments with Text Understanding}
\author {
    Ashim Gupta, 
    Giorgi Kvernadze, 
    Vivek Srikumar\\
}

\affiliations{
    University of Utah\\
    \{\texttt{ashim}, \texttt{giorgi}, \texttt{svivek}\}\texttt{@cs.utah.edu}
}

\begin{document}

\maketitle

\begin{abstract}
In this paper, we study the response of large models from the BERT family to incoherent inputs that should confuse any model that claims to understand natural language. We define simple heuristics to construct such examples. Our experiments show that state-of-the-art models consistently fail to recognize them as ill-formed, and instead produce high confidence predictions on them. 
As a consequence of this phenomenon, models trained on 
sentences with randomly permuted word order
perform close to state-of-the-art models. 
To alleviate these issues, 
we show that if models are explicitly trained to recognize invalid inputs, they can be robust to such attacks without a drop in performance.
\end{abstract}

\section{Introduction}
\label{sec:intro}

The BERT family of  models~\cite[and others]{devlin2019bert,liu2019roberta}
form the backbone of today's NLP systems. At the time of writing, all eleven systems deemed to outperform humans in the GLUE benchmark suite~\cite{wang2018glue} belong to this family. Do these models understand language? Recent work suggests otherwise. For example, \citet{bender2020climbing} point out that models trained to mimic linguistic form (i.e., language models) may be deficient in understanding the meaning conveyed by language.

In this paper, we show that such models struggle even with the form of language by demonstrating that they force meaning onto token sequences devoid of any.
For instance, consider the natural language inference (NLI) example in 
\cref{fig:intro_example}. A RoBERTa-based model that scores $\sim 89\%$ on the Multi-NLI dataset~\cite{williams2018broad} identifies that the premise entails the hypothesis. However, when the words in the hypothesis are  sorted alphabetically (thereby rendering the sequence meaningless), the model still makes the same prediction with high confidence. Indeed, across Multi-NLI, when the hypotheses are sorted alphabetically, the model retains the same prediction in $79\%$ of the cases, with a surprisingly high average confidence of $\sim 95\%$!  
We argue that a reliable model should not be insensitive to such a drastic change in word order.

\begin{figure}[ht]
\tikz\node[draw=black!100!red,inner sep=4pt,line width=0.15mm,rounded corners=0.3cm]{
\begin{tabular}{lp{0.65\columnwidth}}
    Premise & In reviewing this history, it's important to make some crucial distinctions. \\\midrule
Original    & Making certain distinctions is  \\ 
Hypothesis  & imperative in looking back on the past.                                                       \\
            & \hfill \emph{\textsc{Entailment} with probability 0.99}                                                   \\\midrule
Sorted      & back certain distinctions imperative in                                           \\ 
Hypothesis  & is looking making on past the .                                                          \\
            & \hfill \emph{\textsc{Entailment} with probability 0.97}
\end{tabular}
}; 
\caption{An Example for Natural Language Inference from the MNLI (-m)
validation set. For the original premise and hypothesis, RoBERTa fine-tuned model makes the correct prediction that the premise entails the hypothesis. Alphabetically sorting the hypothesis makes it meaningless, but the model still retains the prediction with high confidence.} 
\label{fig:intro_example}
\end{figure}

\begin{table*}[!t]
\centering
\begin{tabular}{C{1.58cm}p{1.7cm}p{10.5cm}p{2.3cm}}
\toprule
\textbf{Dataset}                                                                            & \textbf{Transform} & \textbf{Input}                                                                          & \textbf{Prediction} \\ \midrule
\multirow{5}{*}{\begin{tabular}[c]{@{}c@{}}Natural \\ Language\\ Inference \\ MNLI\end{tabular}}&                         & P:  As with other types of internal controls, this is a cycle of activity, not an exercise with a defined beginning and end. &                     \\\addlinespace
                                                                                                  & Original                        & H:  There is no clear beginning and end, it's a continuous cycle.                   & Ent ($99.48 \%$)           \\
                                                                                                  & \transformation{Shuffled}                &H$_1^\prime$: , beginning end no there clear 's continuous is a it and cycle .                 &       Ent ($99.60 \%$)              \\
                                                                                                  & \transformation{PBSMT-E}                 & H$_2^\prime$:   The relationship of this is not a thing in the beginning .                 &      Ent ($94.82 \%$)               \\
\midrule
\multirow{4}{*}{\begin{tabular}[c]{@{}c@{}}Paraphrase\\Detection\\ QQP\end{tabular}}             &                         & Q1:  How do I find out what operating system I have on my Macbook? &                     \\\addlinespace
                                                                                                  & Original        & Q2:  How do I find out what operating system I have?                   & Yes ($99.53 \%$)          \\
                                                                                                  & \transformation{Repeat}                  & Q2:  out out i find out what out out i find?                  &  Yes   ($99.98 \%$)                \\
                                                                                                  & \transformation{CopySort}                & Q2:   ? do find have how i i macbook my on operating out system what                  &         Yes ($98.52 \%$)           \\ 
\midrule
\multirow{3}{*}{\begin{tabular}[c]{@{}c@{}}Sentiment \\ Analysis\\ SST-2\end{tabular}}          & Original                & A by-the-numbers effort that won't do much to enhance the franchise.                 &                --ve ($99.96 \%$)      \\
                                                                                                  & \transformation{Sort}                    & a by-the-numbers do effort enhance franchise much n't that the to wo.                  &       --ve ($99.92 \%$)               \\
                                                                                                  & \transformation{Drop}                    & a--n won do to franchise.                  &        --ve ($99.96 \%$)             \\  
     
\bottomrule                                                                         \end{tabular}
\caption{We generate invalid token sequences using destructive transformations that render the inputs meaningless. A fine-tuned RoBERTa model assigns a high probability (in parenthesis) to the same label as the original example. 
For NLI, the model chooses between \textit{entail}, \textit{contradict}, and \textit{neutral}. For sentiment analysis, possible labels are --ve or +ve. For paraphrase detection, model answers if the two texts are paraphrases of each other (Yes or No). The appendix contains more such examples.}
\label{fig:examples}
\end{table*}

We study the response of large neural models to \emph{destructive transformations}: perturbations of inputs that render them meaningless. 
 \Cref{fig:intro_example} shows an example. We define several such transformations, all of which erase meaning from the input text and produce token sequences that are not natural language (i.e., word salad).

We characterize the response of models to such transformations using two metrics: its ability to predict valid labels for invalid inputs, and its confidence on these predictions.
Via experiments on three tasks from the GLUE benchmark, we show that the labels predicted by state-of-the-art models for destructively transformed inputs bear high agreement with the original ones. Moreover, the models are highly confident in these predictions. 
We also find that models trained on meaningless examples perform comparably to the original model on unperturbed examples, despite never having encountered any well-formed training examples. Specifically, models trained on meaningless sentences constructed by permuting the word order perform almost as well as the state-of-the-art models. 
These observations suggest that, far from actually understanding natural language, today's state-of-the-art models have trouble even \emph{recognizing} it.

Finally, we evaluate strategies to mitigate these weaknesses using regularization that makes models less confident in their predictions, or by allowing models to reject inputs.

In summary, our contributions are\footnote{Our code is available at \url{https://github.com/utahnlp/word-salad}}: 
\begin{enumerate}
\item We define the notion of destructive input transformations to test the ability of text understanding models at processing word salad. We introduce nine such transformation functions that can be used by practitioners for diagnostic purposes without requiring additional annotated data. 
\item We show via experiments that today's best models force meaning upon invalid inputs; i.e., they are not using the right kind of information to arrive at their predictions. 
\item We show that simple mitigation strategies can teach models to recognize and reject invalid inputs. 
\end{enumerate}

\section{Tasks and Datasets}
\label{sec:tasks-data}

Our goal is to demonstrate that state-of-the-art models based on
the BERT family do not differentiate between valid and
invalid inputs, and that this phenomenon is ubiquitous.  To
illustrate this, we focus on three tasks (\cref{fig:examples})
, which
also serve as running examples.

\textbf{Natural language inference} (NLI) is the task of
determining if a premise entails, contradicts, or is unrelated to a
hypothesis. We use the MNLI~\cite{williams2018broad} and~SNLI \cite{bowman2015large} datasets.

\textbf{Paraphrase detection} involves deciding if two sentences
are paraphrases of each other. For this task,  we use the Microsoft Research
Paraphrase Corpus~\cite[MRPC,][]{dolan2005automatically}, and Quora Question Pair
(QQP) dataset\footnote{\url{https://www.quora.com/q/quoradata/First-Quora-Dataset-Release-Question-Pairs}}.

\textbf{Sentiment classification} requires predicting whether a sentence has a positive or negative sentiment. We use the Stanford Sentiment Treebank~\cite[SST-2,][]{socher2013recursive}.

\section{Destructive Transformations}
\label{sec:transformation-functions}

There has been a growing interest in studying input perturbations~\cite[e.g.,][]{ebrahimi2018hotflip,alzantot2018generating,wallace2019universal,DBLP:conf/aaai/JinJZS20,ren2019generating,DBLP:conf/cikm/Garg0GL20}. Given a model for a task, some input perturbations preserve labels. For example, a true paraphrase of a sentence should not change its sentiment. Certain other perturbations force labels to change in controlled ways. For example, a negated hypothesis should change the label in the NLI task.

In this paper, we focus on a new class of perturbations---\emph{destructive transformations}---which render inputs invalid. Because any informative signal in the input is erased, the transformed examples should not have \emph{any} correct label.\footnote{We refer to such changes to inputs as transformations, instead of perturbations (as in adversarial perturbations), to highlight the fact that change in the inputs need not be small.} 

For example, in the NLI task, a hypothesis whose words are shuffled is (with high probability) not a valid English sentence. The transformed hypothesis cannot contain information to support an \textit{entail} or a \textit{contradict} decision. Moreover, it is not a sentence that is unrelated to the premise---it is not a sentence at all! Indeed, the transformation creates an example that lies outside the scope of the NLI task.

Yet, the premise and transformed hypothesis fit the interface of the problem. This gives us our key observation: While NLP models are typically trained to work on well-formed sentential inputs, they accept any sequence of strings as inputs. Of course, not all token sequences are meaningful sentences; we argue that we can gain insights about models by studying their response to meaningless input sequences  (i.e., \textit{invalid inputs}) that are carefully crafted from valid inputs. One advantage of this protocol of using transformations is that we do not need to collect any new data and can use original training and validation sets.

Let us formally define destructive transformations. Consider a task with input $x \in X$ and an oracle function $f$ that maps inputs to labels $y \in Y$. A destructive transformation $\pi: X \to X$ is a function that operates on $x$ to produce transformed inputs $x^\prime = \pi(x)$ such that $f(x^\prime)$ is \textit{undefined}. That is, none of the labels 
(i.e., the set $Y$) can apply to $x^\prime$.

Destructive transformations can be chained: if $\pi_1$ and $\pi_2$ are destructive transformations for a given input $x$, then $\pi_1(\pi_2(x))$ is also a destructive transformation. We can think of such chaining as combining different diagnostic heuristics.
For tasks whose input is a pair of texts, $x = (x_1, x_2)$, transforming either or both the components should destroy any meaningful signal in the input $x$. For example, in the NLI task, given an input premise and hypothesis, destroying either of them renders the example invalid. For tasks with a pair of texts as input, for our analyses, we only transform one of the inputs, although there is no such requirement.

Next, let us look at different classes of transformations.

\subsection{Lexical Overlap-based Transformations}
\label{sec:lexical-transformations}

These transformation operators preserve the bag-of-words representation of the original input but change the word order. They are designed to diagnose the sensitivity of models to the order of tokens in inputs. \Cref{tab:lexical-overlap-transformations} shows the four lexical-overlap based transformations we define here.

\begin{table}
  \centering
  \begin{tabular}{rp{2.2in}}
    \toprule
    Name              & Description                         \\
    \midrule
    \transformation{Sort}     & Sort the input tokens      \\
    \transformation{Reverse}  & Reverse the token sequence          \\
    \transformation{Shuffle}  & Randomly shuffle tokens         \\
    \transformation{CopySort} &  Copy one of the input texts and then sort it to create the second text. (Only applicable when the input is a pair of texts) \\
    \bottomrule
\end{tabular}
\caption{Lexical-overlap based transformations}
\label{tab:lexical-overlap-transformations}
\end{table}

We ensure that \transformation{Shuffle} sufficiently changes the input by repeatedly shuffling till no bigram from original input is retained. The \transformation{CopySort} operation only applies to tasks that have multiple input texts such as NLI and paraphrase detection. As an example, for the NLI task, given a premise-hypothesis pair, it creates a transformed pair whose hypothesis is the alphabetically sorted premise.
\subsection{Gradient-based Transformations}
\label{sec:gradient-based-transformations}

These transformations seek to study the impact of removing, repeating, and replacing tokens. To decide which tokens to replace, they score input tokens in proportion to their relative contribution to the output. One  (admittedly inefficient) way to compute token importance is to calculate the change in output probability when it is removed. Recent work~\cite[e.g.,][]{ebrahimi2018hotflip,feng2018pathologies} suggests that a gradient-based method is a good enough approximation and is much more efficient. We adopt this strategy here.

Given a trained neural model $\sM$, and the task loss function $\sL$, the change in the loss for the $i^{th}$ input token is approximated by the dot product of its token embedding $\bt_i$ and the gradient of the loss propagated back to the input layer $\nabla_{\bt_i, \sM}\sL$. That is, the $i^{th}$ token is scored by  $\bt_i^\intercal\nabla_{\bt_i,\sM}\sL$.

These token scores approximate the relative importance of a token; a higher score denotes a more important token.
We use the tokens in the \emph{bottom} $r\%$ as per their score---the least important tokens---to define our gradient-based transformations. We use $r=50\%$. \Cref{tab:gradient-transformations} summarizes the transformations that use importance ranking of the tokens. 

\begin{table}
  \centering
  \begin{tabular}{rp{2.4in}}
    \toprule
    Name              & Description                         \\
    \midrule
    \transformation{Drop}        & Drop the least important tokens.        \\
    \transformation{Repeat}      & Replace the least important tokens with one of the most important ones.          \\    
    \transformation{Replace}     & Replace the least important tokens with random tokens from the vocabulary \\
    \transformation{CopyOne} & Copy the most important token from one text as the sole token in the other. (Only applicable when the input is a pair of texts) \\
    \bottomrule
\end{tabular}
\caption{Gradient-based transformations}
\label{tab:gradient-transformations}
\end{table}

\subsection{Statistical transformation: \transformation{PBSMT}}

Recent analyses on the NLI task have shown that neural models rely excessively on shallow heuristics~\cite[]{gururangan2018annotation,poliak2018hypothesis,mccoy2019right}. In particular,~\citet{gururangan2018annotation} showed that annotation artifacts lead to certain words being highly correlated with certain inference classes. For example, in the SNLI data, words such as \textit{animal}, \textit{outdoors} are spuriously correlated with the \textit{entail} label.

Inspired by this observation, we design a transformation scheme that creates invalid examples, and yet exhibit such statistical correlations. We employ a traditional phrase-based statistical machine translation (PBSMT) system to generate examples that use phrasal co-occurrence statistics.

For each label in the task, we train a separate sequence generator that uses co-occurrence statistics for that label. For example, for the NLI task, we have three separate generators, one for each label. Suppose we have a premise-hypothesis pair that is labeled as \textit{entail}. We destroy it using the premise as input to a PBSMT system that is trained only on the entailment pairs in the training set. We use the Moses SMT toolkit~\cite{koehn2007moses} for our experiments.

Why should a system trained to generate a sentence that has a certain label (e.g., an entailment) be a destructive transformer? To see this, note that unlike standard machine translation, we use very limited data for training. Moreover, the language models employed~\cite{heafield2011kenlm} are also trained only on examples of one class. As a result, we found that the produced examples are non-grammatical, and often, out of context. The hypothesis $H_{2}^\prime$ in \cref{fig:examples}, generated using \transformation{PBSMT-E} (i.e., PBSMT for entailments), is one such sequence. We refer to this transformation as \transformation{PBSMT}.

\subsection{Are the Transformations Destructive?}
\label{sec:crowdsource}

To ascertain whether our nine transformations render sentences invalid, we asked crowd workers on Amazon's Mechanical Turk to classify each instance as \textit{valid} or \textit{invalid}, the latter category is defined as an example that is 
incomprehensible, and is therefore meaningless. 

\begin{table}[ht]
  \centering
  \begin{tabular}{lr}
    \toprule
    \textbf{Transformation}      & \textbf{\% Invalid} \\
    \midrule
    Un-transformed               & 7.83                \\
    \midrule
    \transformation{Sort}        & 94.07               \\
    \transformation{Reverse}     & 95.59               \\
    \transformation{Shuffle}     & 94.20               \\
    \transformation{CopySort}    & 95.42               \\
    \midrule
    \textbf{Avg. Lexical}        & 94.82               \\
    \midrule
    \transformation{Replace}     & 91.21               \\
    \transformation{Repeat}      & 100.00              \\
    \transformation{Drop}        & 85.79               \\
    \transformation{CopyOne} & 100.00              \\
    \midrule
    \textbf{Avg. Gradient}       & 94.25               \\
    \midrule
    \transformation{PBSMT}       & 79.92               \\
    \bottomrule
  \end{tabular}
  \caption{Results of crowdsourcing experiments where annotators are asked to labels sentences as meaningful or not.  We aggregate sentence labels from three workers. 
  }
  \label{tab:crowd}
\end{table}

We sampled 100 invalid sentences generated by each transformation (900 in total) and an equal number of sentences from the original (\textit{un-transformed}) validation sets. For each sentence, we collect validity judgments from three crowd workers and use the majority label.~\Cref{tab:crowd} shows the percent of sentences marked as invalid; we see that all the transformations make their inputs incomprehensible.

\section{Measuring Responses to Invalid Inputs}
\label{sec:metrics}

We will now define two metrics to quantify model behavior for invalid inputs. Invalid inputs, by definition, are devoid of information about the label. Consequently, they \emph{do not} have a correct label. If the transformations are truly destructive, a \textit{reliable} model will pick one of the labels at random and would do so with low confidence. That is, a reliable model should exhibit the following behavior: a) the agreement between original predictions and predictions on their transformed invalid variants should be random, and, b) predictions for invalid examples should be uncertain. These expected behaviors motivate the following two metrics.

\noindent \textbf{Agreement} is the \% of examples whose prediction remains same after applying a destructive transformation. A model with agreement closer to random handles invalid examples better.
For the operators designed for tasks with a pair of inputs, namely \transformation{CopySort} and \transformation{CopyOne}, tokens from one of the inputs are copied into another. In such cases, measuring agreement with original input is not useful. Instead, we measure agreement with a default label. For the NLI task, the default label is \textit{entail}, because neural models tend to predict entailment when there is a 
lexical overlap between the premise and hypothesis~\cite{mccoy2019right}. 
Following the same intuition, for paraphrase detection, the default is to predict that the pair is a paraphrase.

\noindent \textbf{Confidence} is defined as the average probability of the predicted label. We want this number to be closer to $\frac{1}{N}$, where $N$ is the number of classes.\footnote{We could alternatively define confidence using the entropy of the output distribution. In our experiments, we found that confidence, as defined here, and entropy reveal the same insights.}

\section{Experiments}
\label{sec:experiments}

\begin{table}
  \centering\
  \begin{tabular}{lcc}
    \toprule
    \textbf{Dataset} & \textbf{Accuracy} & \textbf{Confidence} \\
    \midrule
    SNLI    & 90.87    & 98.38      \\
    MNLI    & 87.31    & 98.27      \\
    QQP     & 90.70    & 98.89      \\
    MRPC    & 89.46    & 98.40      \\
    SST-2   & 94.04    & 99.75      \\
    \bottomrule
  \end{tabular}
  \caption{\textbf{Baseline Performance}, Accuracy and Average Confidence for RoBERTa-\textit{base} on validation sets. 
  For MNLI, we used MNLI-matched for experiments.
  }
  \label{tab:baseline-acc}
\end{table}

For our primary set of experiments, we use RoBERTa~\cite{liu2019roberta} as a representative of the BERT family, whose fine-tuned versions have tended to outperform their BERT counterparts on the GLUE tasks~\cite{dodge2020fine}. 
We use the \textit{base} variant of RoBERTa that is fine-tuned for three epochs across all our experiments, using hyperparameters suggested by the original paper. These models constitute our baseline. \Cref{tab:baseline-acc} shows the accuracy and average confidence of the baseline on the original validation sets.

\subsection{Results and Observations}
\label{sec:results_baseline}

\begin{table}
\setlength{\tabcolsep}{5pt}
  \centering
    \begin{tabular}{l|ccccc}
      \toprule
      \textbf{Transform} & \textbf{MNLI} & {\textbf{SNLI}} & {\textbf{QQP}} & {\textbf{MRPC}} & {\textbf{SST2}} \\

      \midrule
      \transformation{Sort}                  & 79.1         & 82.6           & 88.3          & 81.1          & 83.3            \\
      \transformation{Reverse}               & 76.9        & 75.1         & 86.8         & 77.9           & 82.5            \\
      \transformation{Shuffle}               &  79.4        & 81.1         & 88.4          & 80.4           & 84.8            \\
      \transformation{CopySort}              & 90.5         & 81.3           & 93.5          & 96.8           & --               \\
      \midrule
      \textbf{Avg. Lex.}   & 82.4         & 80.1           & 89.3          & 84.1           & 83.5            \\
      \midrule
      \transformation{Replace}               & 63.0         & 51.9           & 69.9          & 56.6           & 78.1            \\
      \transformation{Repeat}                & 49.7         & 68.5           & 77.1          & 68.1           & 81.3            \\
      \transformation{Drop}                  & 69.4         & 72.7           & 80.4          & 76.7           & 82.5            \\
      \transformation{CopyOne}           & 80.4         & 83.7           & 98.9          & 100             & --               \\
      \midrule
      \textbf{Avg. Grad.}  & 65.6         & 69.2           & 81.6          & 75.4           & 80.6            \\
      \midrule 
      {\transformation{PBSMT}}        & 57.0         & 65.6           &      72.5          & --              & 75.2            \\
            \midrule
      Random                & 33.3         & 33.3           & 50.0          & 50.0           & 50.0            \\
      \bottomrule
    \end{tabular}
  \caption{Agreement scores between predictions from transformed validation set and original validation set. The closer the numbers are to random better the model behavior is.`--` means the transformation is not defined for that dataset. We do not use PBSMT for MRPC as it is a much smaller dataset.}
  \label{table:baseline_transformation_results}
\end{table}

We apply the destructive transformation functions described
earlier
to each task's validation set. To account for the randomness in the \transformation{Shuffle} transformation, its results are averaged across five random seeds. For \transformation{PBSMT} on SST-2, we use the first half sentence as input and train to predict the second half of the sentence. \Cref{table:baseline_transformation_results} shows the agreement results, and~\cref{table:baseline_transformation_confidence} shows average confidence scores.

\paragraph{High Agreement Scores} The high agreement scores show that models retain their original predictions even when label-bearing information is removed from examples. This is a puzzling result: the transformations render sentences meaningless to humans, but the model knows the label. How can the model make sense of these nonsensical inputs? 

We argue that this behavior is not only undesirable but also brings into question the extent to which these models understand text. It is possible that, rather than understanding text, they merely learn spurious correlations in the training data. 
That is, models use the wrong information to arrive at the right answer. 

\paragraph{High Confidence Predictions.} Not only do models retain a large fraction of their predictions, they do so with high confidence (\cref{table:baseline_transformation_confidence}). 
This behavior is also undesirable: a \textit{reliable} model should know what it does not know, and should not fail silently. It should, therefore, exhibit be uncertain on examples that are uninformative about the label.

Research on reliability of predictions suggests that these models are poorly calibrated~\cite{guo2017calibration}. 

\begin{table}
  \centering
    \begin{tabular}{lccccc}
      \toprule
       & \textbf{MNLI} & {\textbf{SNLI}} & {\textbf{QQP}} & {\textbf{MRPC}} & {\textbf{SST-2}} \\
      \midrule
      {Baseline}                  & 94.63         & 92.46           & 98.78          & 97.77           & 99.13            \\
      Random                & 33.33         & 33.33           & 50.00          & 50.00           & 50.00            \\

      \bottomrule
    \end{tabular}
  \caption{Average Confidence over predictions from transformed validation set. We want the the numbers to be closer to random (last row). Refer appendix for full results.}
  \label{table:baseline_transformation_confidence}
\end{table}

\paragraph{Specific transformations.} Invalid examples constructed by lexical transformations are more effective than others, with all agreements over $80\%$. 
Examples from such transformations have high lexical overlap with the original input. 
Our results suggest that models do not use input token positions effectively. We need models that are more sensitive to word order; lexical transformations can be used as a guide without the need for new test sets.

We find that SNLI models have higher agreement scores than MNLI ones for both gradient and statistical correlation based invalid examples. This could mean that the former are more susceptible to gradient-based adversarial attacks. Moreover, the lower scores for \transformation{PBSMT} on the MNLI model shows that it relies less on these statistical clues than SNLI---corroborating an observation by~\citet{gururangan2018annotation}.

\paragraph{Human response to transformed inputs.} Results in~\cref{tab:crowd} show that transformed sentences are invalid. We now perform another set of human experiments to determine if the invalid examples generated (by transformations) make it difficult to perform the classification task 
. This mimics the exact setting that all models are evaluated on by asking humans to perform classification tasks on invalid inputs. Concretely, we ask turkers to perform the NLI task on 450 destructively transformed inputs (50 for each transformation) by ``\textit{reconstructing the inputs to the best of their abilities}''. 
We found that turkers can only ‘predict’ the correct label for invalid examples in 35\% of the cases as opposed to 77\% for original un-transformed examples. These results reinforce the message that large transformer models can make sense of meaningless examples, whereas humans are near-random. 

\subsection{Analysis \& Discussion}
\label{sec:analysis}

\paragraph{Are calibrated models more reliable?}
\label{subsec:calibration}

Neural networks have been shown to produce poorly calibrated probabilities, resulting in high confidence even on incorrect predictions~\cite{guo2017calibration}. Research in computer vision has shown that improving model calibration improves adversarial robustness as well as out-of-distribution detection~\cite{DBLP:conf/iclr/HendrycksG17,thulasidasan2019mixup,hendrycks2019using}. Given the confidence scores in~\cref{table:baseline_transformation_confidence}, a natural question is: \textit{Does improving the calibration of BERT models improve their response to invalid examples? }
We answer this question by training confidence calibrated classifiers using three standard methods.

First we use \textit{label smoothing}, in which training is done on soft labels, with loss function being a weighted average of labels and uniform probability distribution \cite{DBLP:conf/iclr/PereyraTCKH17}. \textit{Focal loss} prevents the model from becoming over-confident on examples where it is already correct. ~\citet{DBLP:conf/nips/MukhotiKSGTD20} showed that focal loss improves calibration of neural models. \textit{Temperature scaling} is a simple calibration method that scales the network's logit values before applying the softmax~\cite{guo2017calibration,DBLP:conf/emnlp/DesaiD20}.

We use \textit{Expected Calibration Error}~\cite[ECE,][]{naeini2015obtaining} to measure a model's calibration error. 
Due to space constraints, we refer the reader to the original work for a formal definition. Better calibrated models have lower ECE. All three methods improve calibration of the original model;~\cref{tab:calibration} shows results on the MNLI validation data. However,~\cref{tab:compare_calibration} shows that none of them improve model response to invalid examples. 

\begin{table}
  \centering
  \begin{tabular}{lcc}
    \toprule
    \textbf{Calibration Method} & \textbf{Accuracy} & \textbf{ECE} \\
    \midrule
    Baseline    & 87.31    & 0.11      \\
    Label Smoothing    & 86.89    & 0.06      \\
    Focal Loss     & 86.98    & 0.05      \\
    Temperature Scaling    & 87.31    & 0.09      \\
    \bottomrule
  \end{tabular}
  \caption{Accuracy on the original validation set and the Expected Calibration Error (ECE) on the validation set for MNLI. 
  Accuracy with temperature scaling is the same as baseline since it is a post-training method for calibration. 
  }
  \label{tab:calibration}
\end{table}

\begin{table}
    \centering
    \begin{tabular}{lcccc}
    \toprule
        & \textbf{B } & \textbf{LS } & \textbf{FL } & \textbf{B + TS}\\
    \midrule
        Lexical &82.35& 81.85& 80.39 & 81.49\\
        Gradient  &60.65& 59.51&  60.32 & 59.18\\
        PBSMT  &57.02& 56.30&  56.49 & 57.04\\
    \bottomrule
    \end{tabular}
    \caption{Average agreement for three calibration methods on MNLI. Calibration does not improve model's response to invalid inputs. B: Baseline, LS: Label Smoothing, FL: Focal Loss, B + TS: Temperature Scaling on baseline.}
    \label{tab:compare_calibration}
\end{table}

\paragraph{Impact of pretraining tasks.}
We now investigate the impact of pre-training tasks on a model's response to invalid examples. 
Both BERT and RoBERTa use a word-based masked language modeling (W-MLM) as the auto-encoding objective. BERT uses Next Sentence Prediction (NSP) as an additional pre-training task. 
We experiment with other BERT variants pre-trained with different tasks: ALBERT~\cite{lan2019albert} uses Sentence Order Prediction (SOP), SpanBERT~\cite{joshi2020spanbert} and BART~\cite{DBLP:conf/acl/LewisLGGMLSZ20} use Span-based MLM (S-MLM) instead of W-MLM. SpanBERT additionally uses NSP, while BART uses a Sentence Shuffling (SS) pretraining objective. ELECTRA~\cite{clark2019electra} uses a Replaced Token Detection (RTD) instead of an MLM objective.

These models are trained on different corpora, and use different pre-training tasks. Despite their differences, the results presented in~\cref{tab:more_models} suggest that all of these models are similar in their responses to invalid examples. These results highlight a potential weakness in our best text understanding systems.

\paragraph{Different Inductive Bias.}
All variants of BERT considered thus far are trained with one of the auto-encoding (AE) objectives and perform rather poorly. This raises a question: \textit{Would models that explicitly inject a word order based inductive bias into the model perform better?} 

To answer this question, we consider three auto-regressive (AR) models with a recurrent inductive bias, namely, ESIM-Glove~\cite{chen2017enhanced}, ESIM-ELMo, and XLNet~\cite{yang2019xlnet}. Both ESIM models are LSTM based models, while XLNet is a transformer-based model that is trained using an auto-regressive language modeling objective along with Permutation LM (P-LM). ESIM-Glove does not use any other pre-training task, while ESIM-ELMo is based on ELMo~\cite{peters2018deep} which is pre-trained as a traditional auto-regressive LM. 

The results are shown in~\cref{tab:more_models}. Again, the results are similar to models trained with auto-encoding objective.
Surprisingly, even a strong recurrent inductive bias is unable to make the models sensitive to the order of words in their inputs: all the AR models have high agreement scores (over $75\%$) on lexical overlap-based transformations. We refer the reader to the appendix for more results.

\paragraph{Bigger is not always better.}
\label{subsec: model_size}
While larger BERT-like models show better performance~\cite{devlin2019bert,JMLR:v21:20-074}, we find that same does \textit{not} hold for their response on invalid examples.~\Cref{tab:more_models} shows that larger BERT models (Large vs Base) do not improve response to invalid examples (recall that smaller agreement scores are better).
%
We see that both BERT variants outperform the RoBERTa counterparts; BERT-\textit{base} provides over $4.5\%$ improvement over RoBERTa-\textit{base} in terms of agreement on invalid examples.



\begin{table}[ht]
  \centering
  \begin{tabular}{lllr}
    \toprule
   \textbf{Class} & \textbf{Pretraining} & \textbf{Model} & \textbf{Agreement} \\
    \midrule
    AE            & W-MLM + NSP          & BERT-B         & 67.1               \\
                  &                      & BERT-L         & 69.0               \\
                  & W-MLM                & RoBERTa-B      & 71.7               \\
                  &                      & RoBERTa-L      & 73.5               \\
                  & W-MLM + SOP          & ALBERT-B       & 67.6               \\
    \midrule                      
    AE            & S-MLM + NSP          & SpanBERT-B     & 67.6               \\
                  & S/W-MLM + SS         & BART-B         & 70.0               \\
                  & RTD                  & ELECTRA-B      & 68.8               \\
    \midrule                      
    AR            & P-LM                 & XLNet-B        & 70.1               \\
                  & LM                   & ESIM- ELMo     & 75.8               \\    
                  & -                    & ESIM- Glove    & 73.5               \\

    \bottomrule
  \end{tabular}
  \caption{Agreement score of different models on MNLI. \textbf{B} refers to the base variant, \textbf{L} refers to the large one. \textbf{AE} refers to models pretrained with auto-encoding objective, \textbf{AR} refers to auto-regressive models. Refer text for full key.}
  \label{tab:more_models}
\end{table}

\paragraph{Small vs. large perturbations.}
Previous work on adversarial robustness  ~\cite[]{alzantot2018generating,DBLP:conf/aaai/JinJZS20,ebrahimi2018hotflip} suggests that robustness of the model to small input perturbations is desirable, meaning that a model's prediction should not change for small perturbations in the input. However, excessive invariance to large input perturbations is undesirable~\cite{DBLP:conf/iclr/JacobsenBZB19}.
Our focus is not on small input changes, rather large ones that destroy useful signals (i.e., destructive transformations). The three types of transformations we discuss in this work achieve this in different ways. We argue that language understanding systems should not only provide robustness against small perturbations (adversarial robustness) but also recognize and reject large perturbations (studied in this work).

\paragraph{Are models learning spurious correlations?}
\label{subsec:train_on_invalid}
The results presented in this work raise an important question: \emph{Why does this undesirable model behavior occur in all models, irrespective of the pretraining tasks, and is even seen in models with a recurrent inductive bias?} We hypothesize that this behavior occurs because these large models learn spurious correlations present in the training datasets, studied previously by~\citet{gururangan2018annotation,DBLP:conf/acl/MinMDPL20}. A simple experiment substantiates this claim. So far, we trained models on valid data and evaluated them on both valid and invalid examples. We now flip this setting: we train on \emph{invalid} inputs generated by a transformation and evaluate on well-formed examples from the validation sets. 

\begin{table}
    \centering
    \begin{tabular}{lccccc}
    \toprule
        & \textbf{MNLI} & \textbf{SNLI} & \textbf{QQP} & \textbf{MRPC} & \textbf{SST-2} \\
    \midrule
        Shuffled &84.56& 89.44& 92.15 & 84.80 & 91.97\\
        Original  &87.31& 90.70&  94.04 & 89.46 & 94.04\\
    \bottomrule
    \end{tabular}
    \caption{Training on only invalid examples generated from \transformation{Shuffle}, evaluation is on original validation data.}
    \label{tab:invalid_train}
\end{table}
\Cref{tab:invalid_train} presents accuracies on the original validation examples for five datasets. We observe that models trained only on shuffled examples perform nearly as well (within $97\%$ for MNLI) as the ones trained on valid examples (second row)!
These observations demonstrate that our models do not use the right kind of evidence for their predictions, or at least, the kind of evidence a human would use. 
This result should raise further concerns about whether we have made real progress on language understanding. 







\section{Mitigation Strategies}
\label{sec:mitigation}

We evaluate three mitigation strategies to alleviate the problem of high certainty on invalid inputs. The goal is to give models the capability to recognize invalid inputs. Two strategies augment the training data with invalid examples. All three introduce new hyperparameters, which are tuned on a validation set constructed by sampling $10 \%$ of the training set. 
The final models are then trained on the full training set.

\paragraph{Entropic Regularization}
\label{sec:entropic-regularization}
The central problem that we have is that the models have high certainty on invalid inputs. To directly alleviate the issue, we can explicitly train them to be less certain on invalid inputs. We do so by augmenting the loss function with a new term. Let $D$ be the original dataset and $D^{'}$ be its complementary invalid dataset. The new training objective is then defined as
\begin{equation}
    L(\text{model}) = L_D(\text{model}) + \lambda H_{D^{'}}(\text{model})
    \label{eq:entropy}
\end{equation}
where $L_{D}$ is the standard cross-entropy loss, and $H_{D^{'}}$ denotes the entropy of model probabilities over invalid examples. The hyperparameter $\lambda$ weighs the relative impact of the two terms. \citet{feng2018pathologies} used similar entropic regularization to improve interpretability of neural models. 

\begin{table}
    \centering
    \begin{tabular}{lccc}
    \toprule
        &{\bf B + Th}&{\bf Ent + Th}&{\bf B + Invalid}\\
    \midrule
        MNLI & 83.40/ 57.95 &\textbf{86.11}/ 89.22  &85.44/ \textbf{97.10} \\
        SNLI  & 88.01/ 54.68 & 89.65/ 93.54 & \textbf{90.88}/ \textbf{98.41}\\
        QQP & 90.25 / 29.20 & \textbf{90.29} / 88.72 & 90.08 / \textbf{95.24}\\
        MRPC &89.46 / 36.82 & 88.24 / 99.43 & \textbf{88.73} / \textbf{99.78} \\
        SST-2  & 90.37 / 35.79 & 92.66 /  95.41 & \textbf{92.78} / \textbf{96.35}\\
    \bottomrule
    \end{tabular}
    \caption{\textbf{Comparison of mitigation strategies}. First number in each cell is accuracy on original validation set. Second number is the \% of examples correctly classified as invalid. Test set for invalid contains examples generated with all nine transformation functions. \textbf{B} refers to the baseline model.
    }
    \label{tab:reject}
\end{table}

We initialize the model with fine-tuned weights from the baseline model and train for three more epochs with the new training objective. The appendix provides further details.

\paragraph{Thresholding Model Probabilities}
\label{sec:thresholding-model-prob}
From our results in~\cref{table:baseline_transformation_confidence}, we observe that although models are confident on invalid examples, their confidence is higher on valid ones. Following this, we experiment with a straightforward approach that thresholds output probabilities to tell valid and invalid examples apart. We used temperature scaling to ensure that the classifier probabilities are calibrated.

This approach is parameterized by a threshold $\theta$: if the maximum probability of the classifier's output is below $\theta$, we deem the input invalid. We used grid search in the range $[\frac{1}{N}, 1.0]$ to find the best performing $\theta$ on a separate validation set. Here, $N$ represents the number of labels.



\paragraph{Invalid as an extra class (B + Invalid)}
\label{sec:learning-reject}
Since one of the goals is to be able to recognize invalid inputs, we can explicitly introduce a new class, \textit{invalid}, to our classification task. The training objective for this new $N+1$ class classification task remains the same, i.e. cross-entropy loss. 

\subsection{Results}
\label{sec:mitigation-results}

With entropic regularization, we observe a significant drop in agreement scores on invalid examples. Indeed, the agreement scores on invalid examples decrease to an average of $35\%$ after regularization. We also notice a significant increase in uncertainty on invalid examples. However, we see that in some cases, accuracies on the original validation set drop by over $1\%$, suggesting a trade-off between accuracy on valid examples and reliable response on invalid examples.

A well-behaved model should maintain high accuracy on valid data and also reject invalid inputs. To compare the three mitigation strategies on an equal footing, we measure accuracy on the original validation data and the percentage of invalid examples correctly identified.  
Since entropic regularization does not explicitly recognize invalid inputs, we apply the thresholding strategy to it (\textbf{Ent + Th}).



\Cref{tab:reject} compares the three methods. We see that simple thresholding \textbf{(B + Th)} does not work well and having the model learn from invalid examples is beneficial. 
It appears that, out of the three methods, training with an extra \textit{invalid} label best maintains the balance between accuracy and invalid input detection.

We also studied 
if mitigating one kind of transformation helps against others. Using (B + Invalid) we train on one transformation and test on the rest. We found that mitigation can be transferable. The appendix provides detailed results.






\section{Final Words}
\label{sec:discussion}

The main message of this paper is that today's state-of-the-art models for text understanding have difficulties in telling the difference between valid and invalid text. This observation is congruent with several other recent lines of work that highlight the deficiencies of today's text understanding systems. For example,~\citet{feng2018pathologies} construct irregular examples by successively removing words without affecting a neural model's predictions. Adversarial attacks on NLP models~\cite[e.g.][]{DBLP:conf/aaai/JinJZS20} expose their vulnerabilities; for example, \citet{wallace2019universal} offer an illustrative example where a model's prediction can be arbitrarily changed.

Statistical models need not always get well-formed inputs. Consequently, when models are deployed, they should be guarded against invalid inputs, not just in NLP, but also beyond~\cite[e.g.,][]{liang2018enhancing}. 
~\citet{Krishna2020Thieves} showed that it is possible to steal BERT-like models by using their predictions on meaningless inputs. Our work can be seen as highlighting why this might be possible: model predictions are largely not affected even if we destructively transform inputs. Our work shows simple mitigation strategies that could become a part of the standard modeling workflow.

\section{Ethics Statement}
\label{sec:ethics}
Our work
points to a major shortcoming of the BERT family of models: they have a hard time recognizing ill-formed inputs. This observation may be used to construct targeted attacks on trained models, especially publicly available ones. We call for a broader awareness of such vulnerabilities among NLP practitioners, and recommend that NLP models should be actively equipped with mitigations against such attacks. 

\section*{Acknowledgments}
We thank members of the Utah NLP group for their valuable insights, and reviewers for their helpful feedback. This material is based upon work supported by NSF under grants \#1801446 (SATC) and \#1822877 (Cyberlearning).

\bibliography{references}
\appendix

\clearpage
\section{More Examples}

\begin{table*}[!t]
\centering
\resizebox{\textwidth}{!}{
\begin{tabular}{C{1.6cm}p{1.7cm}p{10.5cm}p{2.3cm}}
\toprule
\textbf{Dataset}                                                                            & \textbf{Transform} & \textbf{Input}                                                                          & \textbf{Prediction} \\ \midrule
\multirow{8}{*}{\begin{tabular}[c]{@{}c@{}}Natural \\ Language\\ Inference - \\ MNLI\end{tabular}}&                         & P:  I feel that you probably underestimate the danger, and therefore warn you again that I can promise you no protection. &                     \\\addlinespace
                                                                                                  & Original                        & H:  I warn you again, that I can promise you no protection, as I feel that you probably underestimate the danger.                  & Ent ($99.65 \%$)           \\
                                                                                                  & Sorted                &H$^\prime$: , , again as can danger feel i i i no probably promise protection that that the underestimate warn you you you .                 &       Ent ($98.38 \%$)              \\
                                                                                                   
                                                                                                  \cmidrule{2-4}
                                                                                                   
                                                                                                  &  & P:  However, the specific approaches to executing those principles tended to differ among the various sectors. &                     \\\addlinespace
                                                                                                    
                                                                                                    & Original                        & H:  Specific approaches to each principle is the same in each sector.                  & Con ($99.96 \%$)           \\
                                                                                                  & Reversed                &H$^\prime$: sector each in same the is principle each to approaches specific .                &       Con ($99.94 \%$)              \\
                                                                                                   
                                                                                                  \cmidrule{2-4}
                                                                                                   
                                                                                                  &  & P:   Nash showed up for an MIT New Year's Eve party clad only in a diaper. &                     \\\addlinespace
                                                                                                    
                                                                                                    & Original                        & H: Nash had too many nasty pictures on Instagram.                  & Neu ($99.84 \%$)           \\
                                                                                                  & Repeat                &H$^\prime$: too had had too many many inst inst..                &       Neu ($99.84 \%$)              \\
                                                                                                   
                                                                                                  \cmidrule{2-4}
                                                                                                   
                                                                                                  &  & P:   The number of steps built down into the interior means that it is unsuitable for the infirm or those with heart problems. &                     \\\addlinespace
                                                                                                    
                                                                                                    & Original                        & H:    The interior is well suited for those with cardiac issues.                & Con (99.43 \%)           \\
                                                                                                  & PBSMT-C                &H$^\prime$: there was no way to the is unsuitable for the park and infirm .               &       Con (99.71 \%)              \\
                                                                                                   
\midrule
\multirow{4}{*}{\begin{tabular}[c]{@{}c@{}}Paraphrase\\ Detection - \\ QQP\end{tabular}}             &                         & Q1:  How long can I keep a BigMac in my fridge before eating it? &                     \\\addlinespace
                                                                                                  & Original        & Q2:  How long do refried beans last in the fridge after you open the can?                   & No (99.98 \%)          \\
                                                                                                  & Repeat                  & Q2$^\prime$:  how longDonald cluesried beans International in the fridge111 dessert open the Mass squ                &  No   (99.97 \%)                     \\
                                                                                                   
                                                                                                  \cmidrule{2-4}
                                                                                                   
                                                                                                    &                         & Q1:  How do I post a question in quora? &                     \\\addlinespace
                                                                                                  & Original        & Q2:  How can I ask my question on Quora?                  & Yes (99.97 \%)          \\
                                                                                                  & CopyOne                  & Q2$^\prime$:  ora              &  Yes   (99.98 \%)                     \\

                                                                                                    \cmidrule{2-4}
                                                                                                   
                                                                                                    &                         & Q1:  What is the best programming language to learn first and why? &                     \\\addlinespace
                                                                                                  & Original        & Q2:  What programming language is best (easiest) to learn first?                   & Yes (99.43 \%)          \\
                                                                                                  & Shuffle                  & Q2$^\prime$:  ( first programming best language learn is easiest what ? ) to                &  Yes   (99.41 \%)                     \\

\midrule
\multirow{3}{*}{\begin{tabular}[c]{@{}c@{}}Sentiment \\ Analysis - \\ SST-2\end{tabular}}          & Original                & Directed in a paint-by-numbers manner .                 &                --ve (99.95 \%)      \\\addlinespace
                                                                                                  & Repeat                    &  directed in a a-by-n in inby                  &       --ve (98.45 \%)               \\
     
                                                                                                \cmidrule{2-4}
                                                                                                & Original                & old-form moviemaking at its best .                 &                +ve (99.98 \%)      \\\addlinespace
                                                                                                  & Replace                    &  CBS scare 1966 moviem GS at<200b> NZ add                 &       +ve (99.98 \%)               \\
     
\bottomrule                                                                     \end{tabular}}
\caption{More Examples for invalid examples generated using destructive transformations that render the inputs meaningless. A fine-tuned RoBERTa model makes the same predictions with very high probability (in parenthesis). \transformation{PBSMT-C} uses a generation model to learn statistical correlations for class Contradict. For NLI, the model chooses between \textit{entail (Ent)}, \textit{contradict (Con)}, and \textit{neutral (Neu)}. For sentiment analysis, possible labels are --ve or +ve. For paraphrase detection, model answers if the two texts are paraphrases of each other (Yes or No). Notice that in some of the examples, tokens are just the sub-words from the RoBERTa tokenizer. }
\label{fig:examples_appendix}
\end{table*}

\Cref{fig:examples_appendix} shows more invalid examples generated using our transformations.

\section{Additional Technical Details and Observations}
The following list contains some of the technical details and observations not mentioned in the paper:

\begin{enumerate}
    \item For all the lexical overlap transformations, we ensure that the punctuation at the end of sentence remains at the end.
    
    \item We found that some of the datasets had a formatting problem. For example, one example in QQP validation set only had one input. We exclude such examples from our data. 
    
    \item We found that for PBSMT models created for NLI task, agreement scores were much higher for classes: entailment, and contradiction, than neutral class.
\end{enumerate}

\section{Additional Details for Crowd-Sourcing Experiments }
We collected crowd annotations for ascertaining the (in)validity of examples generated by our transformation functions. Here, we discuss the instructions we provided to annotators along with costs. We also briefly describe the quality control procedure we used.

\paragraph{Crowd Workers.} We collected annotations from three Amazon's MTurk workers. We used the majority label among them for our analysis. They were asked to classify sentence as invalid only \textit{if the text is not natural and text is incomprehensible}. 

Every worker was paid \$0.05 for each Human Intelligence Task (HIT), and every HIT consisted of 5 examples.

To ensure that annotation quality, in every HIT, we insert an example which was already annotated (by graduate students). We reject all the annotations from a HIT if worker provides a wrong answer to that example. In all, we only rejected 3\% of the total hits. Because we rejected some annotations, the increments are not in 0.33\% multiples. 

A total of 1800 sentences were annotated in the first human experiment. This had equal number of valid and invalid sentences. This was to ensure that valid and invalid sentences have equal representation and the estimate we get from crowd sourced experiment is un-biased.

\subsection{Human response to classification on invalid examples }
For ascertaining the validity of generated examples from our transformations, first we performed a crowd-sourcing experiment where humans were asked to rate sentences as valid or invalid. We describe an additional set of experiments where humans are asked to perform classification on invalid examples, a setting identical to the one our models work with. We perform this experiment to understand the contrast between humans and models w.r.t their response when presented with invalid examples.

Specifically, for every transformation we study, we sample 50 invalid examples generated (450 total) and ask turkers to perform classification task on these examples. Additionally, we have 50 clean examples to measure their performance on valid examples. For every example, we collect annotations from three annotators and use the majority label for our analyses. Since, invalid examples are expected to be incomprehensible, we provide an additional instruction to the turkers to perform the task by \textit{"reconstructing the inputs to the best of their abilities"}. Rest of the instructions are taken from the original task instructions. 

We perform this analyses on two tasks: NLI (MNLI), and Sentiment Analysis (SST2). We use NLI as a task representative of tasks that require a pair of inputs, and use SST2 as representative of tasks that require only one input. In the main paper, we have only presented these results for MNLI due to space constraints. Here we describe results on both tasks. 

Results of crowd-sourced experiments are presented in~\cref{tab:human_3}. We note that accuracy for humans on invalid examples is much lower than their accuracy on valid (un-transformed) examples. This shows that our proposed transformations in fact, do destroy the label determining information in the inputs. On invalid examples, for MNLI the humans perform with 35\% accuracy, while on SST2 the accuracy is around 49\%. Both of these numbers are near-random: 33\% for MNLI, 50\% for SST2.

\begin{table}[]
\centering
\begin{tabular}{l|c|c}
\toprule
Dataset & Valid &  Invalid \\ 
\midrule
MNLI    & 77\% &    35\%  \\
SST-2 &   83\% & 49\%     \\ 
\bottomrule
\end{tabular}
\caption{Human Accuracy for Valid and Invalid examples on two datasets. Invalid examples are generated from our proposed transformations, while valid examples come from the original validation sets.}
\label{tab:human_3}
\end{table}

\section{Additional Experimental Results }

\subsection{RoBERTa is highly confident: Detailed results}
\label{sec:additional_confidence}
In results section of the main paper, we show that fine-tuned RoBERTa model achieves a high agreement score with high confidence on invalid examples generated using our proposed transformations. We provided average confidence values for the model on each dataset. Here we provide detailed results for all datasets on all transformations. These results are shown in \cref{table:baseline_transformation_confidence_appendix}. 

\begin{table}[htbp]
  \setlength{\tabcolsep}{5pt}
  \centering
    \begin{tabular}{l|ccccc}
      \toprule
      \textbf{Transform} & \textbf{MNLI} & {\textbf{SNLI}} & {\textbf{QQP}} & {\textbf{MRPC}} & {\textbf{SST2}} \\
      \midrule
      Random                & 33.33         & 33.33           & 50.00          & 50.00           & 50.00            \\
      \midrule
      {Sort}                  & 94.86         & 93.17           & 98.76          & 98.19           & 99.33            \\
      {Reverse}               & 94.15         & 89.15           & 98.85          & 97.65           & 99.25            \\
      {Shuffle}               & 94.94         & 92.53           & 98.77          & 97.67           & 99.17            \\
      {CopySort}              & 97.96         & 92.48           & 98.16          & 99.55           & -               \\
      \midrule
      \textbf{Avg. Lex.}   & 94.48         & 91.83           & 98.63          & 98.27           & 99.25            \\
      \midrule
      {Replace}               & 94.81         & 93.35           & 99.56          & 97.62           & 99.21            \\
      {Repeat}                & 90.46         & 91.29           & 99.31          & 94.07           & 99.26            \\
      {Drop}                  & 94.01         & 93.96           & 99.20          & 97.31           & 99.32            \\
      {CopyOne}           & 94.94         & 92.69           & 99.86          & 99.76           & --               \\
      \midrule
      \textbf{Avg. Grad.}  & 93.55         & 92.82          & 99.49          &  97.19          & 99.26            \\
      \midrule 
      {\textbf{PBSMT}}        & 95.05         & 93.52           &   96.56             & -              &    98.39         \\
      \bottomrule
    \end{tabular}
  \caption{Average Confidence over predictions from transformed validation set .  The closer the numbers are to random, better the model. Cells marked `--' indicate that the transformation is not defined for that dataset. }
  \label{table:baseline_transformation_confidence_appendix}
\end{table}

Because MRPC has less than 4000 training examples, we could not generate PBSMT based invalid samples.

\subsection{Comparing models from the BERT-family}
\label{sec:additional_size_results}
We now provide detailed results for four models from the BERT-family on MNLI. We aim to compare these models for their response on invalid inputs generated using our proposed transformation functions. Agreement scores are provided in \cref{table:detailed_size_agreement_appendix}. Confidence scores are provided in \cref{table:detailed_size_confidence_appendix}

\subsection{Mitigation results}
\label{sec:appendix_entropy}



To augment the training data with invalid examples, we sample $50 \%$ examples from training set and for each of them, generate all nine types of invalid examples as described in section on Destructive Transformations. These examples are then augmented to the original training set to get our final augmented training set. We initialize the model with the fine-tuned weights from the baseline model and then train it further for three more epochs with the loss function of~\cref{eq:entropy}. 

\begin{figure}[ht]
    \centering
        \includegraphics[width=\columnwidth]{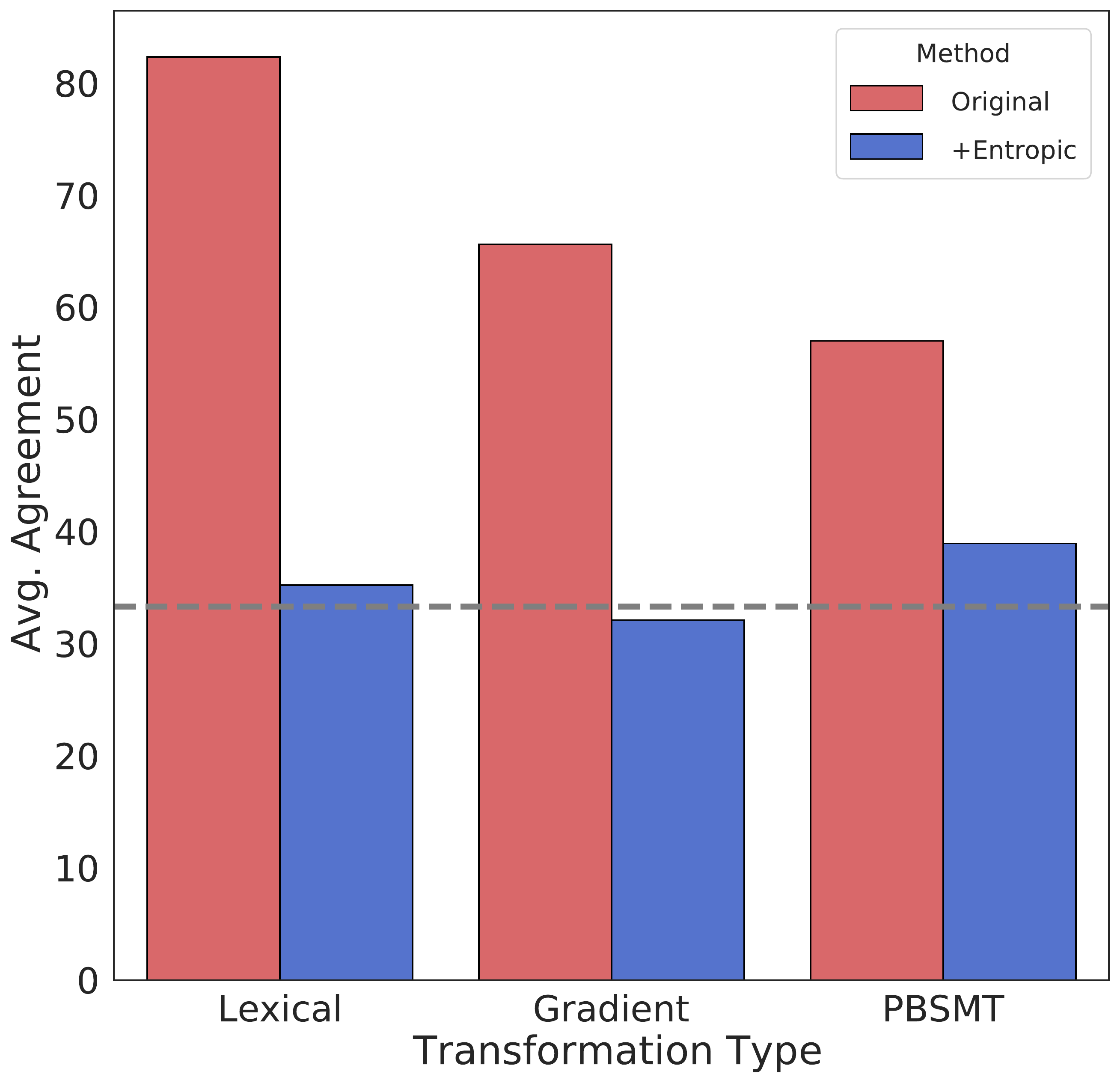}
    \caption{Results after applying entropic regularization. We found that not only did the agreement scores come down to an average of 35\%, average confidence also decreased significantly. Gray line in the figure indicates agreement scores for random predictions.}
    \label{fig:entropy_mnli_appendex}
\end{figure}

After applying entropic regularization, we observe a significant drop in model's agreement scores on invalid examples (refer~\cref{fig:entropy_mnli_appendex}). Indeed, the agreement scores on invalid examples have decreased to an average of 35\% after regularization. Moreover, owing to the regularization, we notice a significant increase in uncertainty on nonsensical examples. However, we also notice that in some cases accuracies on original validation set decrease by more than $1 \%$ . For instance, for MNLI, accuracy decreased from 87.31 to 86.11. Clearly there is some trade-off between accuracy on clean examples and reliable response on invalid examples.

\begin{table*}[ht]
  \centering
    \begin{tabular}{l|cccc}
      \toprule
      \textbf{Transformation} & \textbf{BERT-\textit{base}} & {\textbf{BERT-\textit{large}}} & {\textbf{RoBERTa-\textit{base}}} & {\textbf{RoBERTa-\textit{large}}}\\
     \midrule
      {Sort}                  & 76.97         & 78.54           & 79.19          & 81.20             \\
      {Reverse}               & 74.32         & 74.19           & 76.91          & 75.54              \\
      {Shuffle}               & 76.94         & 78.22           & 79.35          & 81.93               \\
      {CopySort}              & 81.49         & 88.86           & 90.48          & 90.48             \\
      \midrule
      \textbf{Avg. Lexical}   & 80.48         & 79.95           & 82.35          & 82.29                \\
      \midrule
      {Replace}               & 56.50         & 59.82           & 63.01          & 64.19                 \\
      {Repeat}                & 45.23         & 49.22           & 49.71          & 49.98               \\
      {Drop}                  & 61.80         & 64.10           & 69.43          &  71.45              \\
      {CopyOne}           & 74.58         &   71.22         & 80.41          & 89.75                     \\
      \midrule
      \textbf{Avg. Gradient}  &  59.37        & 61.09          & 65.63          &  68.84                  \\
      \midrule 
      {\textbf{PBSMT}}        & 56.15         & 57.37           &   57.02             & 57.16                \\
            \midrule
      Random                & 33.33         & 33.33           & 33.33          & 33.33                  \\

      \bottomrule
    \end{tabular}
  \caption{Results comparing average agreement among model sizes. Closer the numbers are to random, better is the model.`--` means the transformation is not defined for that dataset. }
  \label{table:detailed_size_agreement_appendix}
\end{table*}

\begin{table*}[ht]
  \centering
    \begin{tabular}{l|cccc}
      \toprule
      \textbf{Transformation} & \textbf{BERT-\textit{base}} & {\textbf{BERT-\textit{large}}} & {\textbf{RoBERTa-\textit{base}}} & {\textbf{RoBERTa-\textit{large}}}\\
     \midrule
Sort                 & 94.26     & 94.88      & 94.86        & 94.87         \\
Reverse              & 93.45     & 93.8       & 94.15        & 94.39         \\
Shuffle              & 94.31     & 94.97      & 94.94        & 94.03         \\
CopySort             & 94.31     & 94.36      & 97.96        & 96.56         \\
\midrule
\textbf{Avg. Lexical}         & 94.08     & 94.50      & 95.48        & 94.96         \\
Replace              & 89.66     & 91.34      & 94.81        & 94.68         \\
Repeat               & 92.65     & 78.97      & 90.46        & 92.99         \\
Drop                 & 79.19     & 93.39      & 94.01        & 94.71         \\
CopyOne              & 98.49     & 98.63      & 94.94        & 91.45         \\
\midrule
\textbf{Avg. Gradient}        & 89.99     & 90.58      & 93.55        & 93.46         \\
PBSMT                & 94.59     & 95.15      & 95.05        & 95.11         \\ \bottomrule
\end{tabular}
  \caption{Results comparing confidence scores among model sizes. Closer the numbers are to random, better is the model.`--` means the transformation is not defined for that dataset. }
  \label{table:detailed_size_confidence_appendix}
\end{table*}

\section{Training Details}
We will release the code to reproduce all over experiments after acceptance. 

\subsection{Mitigation Strategies}
We used three mitigation strategies in our paper: Thresholding original model probabilities (B + Th), Thresholding Entropic Regularized Model's probabilities (Ent + Th), and training with an extra invalid class (IC). Both (Ent + Th) and IC need invalid examples at training time. For these two methods, we augmented the training data with invalid examples generated by applying all nine types of destructive transformations to randomly selected set of $50\%$ examples from training set. These examples are then combined with the original training set to get our final training set. We initialize the model with the fine-tuned weights from the baseline model and then train it further for two more epochs. 

Since all our experiments are on validation datasets, we cannot use those for searching hyperparameters. To find best hyperparameters, we sample 10 \% examples from training set and use this set for validation. After finding best hyperparameters, the models are trained on full training sets. 

\paragraph{Finding best thresholds} We used two metrics for comparing the three mitigation strategies -- (1) Accuracy on original validation set (Acc.), (2) Percentage of invalid examples identified correctly (\%Invalid). For second metric, we create a set consisting of equal number of invalid examples generated by each transformation. As noted in the section on mitigation strategies in our main paper, there is a trade-off between Acc. and \%Invalid. This is because these models might incorrectly classify examples from original (clean) validation set as invalid. For thresholding both baseline and entropic model, we have to find a balance between the two metrics, which leads to decrease in accuracy on original validation set. 

We follow a simple procedure to find these thresholds using the validation set sampled from training set. We perform a grid search in range $[\frac{1}{N}, 1.0]$ with step size of 0.001, where $N$ is the number of classes for the task. After this, we select all the thresholds that provide accuracy on (clean) validation set within a certain tolerance value (3\%) of the original accuracy. Out of all these thresholds, the threshold that provides best \% Invalid is selected. For example, if accuracy of the Entropic model on MNLI is 87.31, we calculate two metrics for all thresholds between $[\frac{1}{3}, 1.0]$. We then select all thresholds for which clean accuracy is within 3\% of 87.31 (--84.31). From these values, the threshold that provides best detection of invalid examples is selected for analysis.

\paragraph{Regularization Parameter for Entropic Regularization}
Entropic regularization introduces a new parameter $\lambda$ that provides a trade-off between regularization and original cross entropy loss (refer section on Entropic Regularization in main paper). We found that large values of $\lambda$ tend to decrease accuracy on original validation set. We fix $\lambda = 0.1$ for all our experiments. We did try other values in $\{0.01, 0.1, 0.3, 0.5, 1.0, 5.0\}$. 

\paragraph{Training a PBSMT Model}

As described in main paper, we train separate sequence generation models for each label. We found the examples to be non-grammatical as well as completely out of context. With default PBSMT parameters, the model mostly duplicated inputs at its outputs, while adjusting distortion coefficient and language model coefficient achieved our desirable results. 

\paragraph{Issues with training on Gradient based Transformations}
When mitigating with invalid examples generated using gradient based methods, we noticed that sometimes if the number of gradient based invalid examples were large, the model would stop learning and start predicting the same class for every example. To overcome this behavior, simply using less number of gradient based examples worked. For mitigation, we tried using 50\%, or 40\%,  or 30\% of training examples to add to the augmented training set.

\section{Transferability of Mitigation Strategies}
\label{sec:appendix_transfer}
\begin{figure}[ht]
    \centering
 \includegraphics[width=\columnwidth]{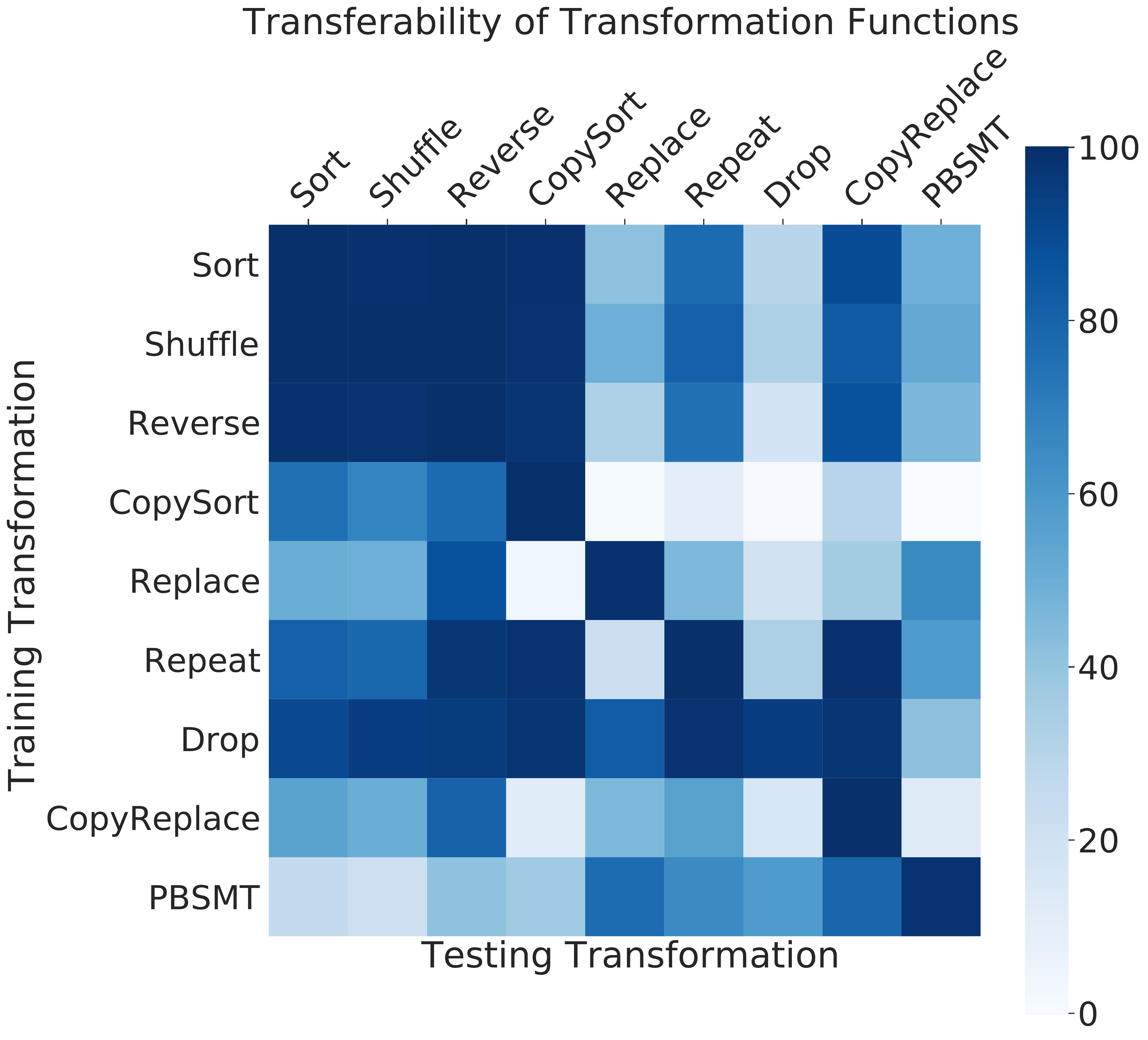}
\caption{Transferability across transformation functions. We notice a great deal of transferability across lexical overlap transformations.}
\label{fig:heatmap_reject}
\end{figure}

We study if mitigation against one transformation helps against others using the strategy of training with the \textit{invalid} label. We train models with invalid examples of one transformation and evaluate on invalid examples of all other transformations. The results in \cref{fig:heatmap_reject}  suggest that  mitigation against transformations within a class type are transferable. For instance, training with any of the \transformation{Sort}, \transformation{Shuffle}, \transformation{Reverse} mitigates other transformations from the same group. However, to mitigate \transformation{Drop} and \transformation{PBSMT}, we need to train on invalid examples of these types. \Cref{table:detailed_reject_transferable_appendix} shows all of the results. 

\begin{table*}[ht]
\centering
\begin{tabular}{l|ccccccccc}
      \toprule
 & Sort  & Reverse & Shuffle & CopySort & Replace & Repeat & Drop  & CopyOne & PBSMT \\ \midrule
Sort                 & 99.66 & 99.82   & 99.43   & 99.39    & 41.45   & 77.19  & 29.33 & 89.2    & 49.25 \\
Reverse              & 99.03 & 99.86   & 98.64   & 97.8     & 32.66   & 74.62  & 17.86 & 87.12   & 45.79 \\
Shuffle              & 99.64 & 99.85   & 99.62   & 98.81    & 49.45   & 81.02  & 32.87 & 83.53   & 52.25 \\
CopySort             & 75.27 & 77.66   & 67.81   & 99.97    & 1.01    & 10.28  & 0.68  & 29.92   & 0.05  \\
Replace              & 50.21 & 87.13   & 49.22   & 3.69     & 99.11   & 45.29  & 19.26 & 36.25   & 65.22 \\
Repeat               & 80.97 & 96.93   & 78.88   & 98.61    & 22.05   & 99.93  & 33    & 99.47   & 58.25 \\
Drop                 & 90.61 & 95.22   & 94.42   & 97.74    & 82.9    & 98.73  & 94.81 & 97.62   & 42.02 \\
CopyOne              & 55.26 & 80.15   & 50.14   & 11.45    & 45.47   & 55.77  & 15.93 & 99.83   & 12.68 \\
PBSMT                & 25.45 & 41.25   & 21.26   & 36.5     & 76.56   & 65.2   & 58.56 & 79.25   & 98.64 \\ 
\bottomrule
\end{tabular}
 \caption{Results on transfer effects of each transformation function. The labels on the left indicate the function that the model was trained on, labels on the top show the functions it was tested on. All numbers denote percentage of invalid examples detected. We note that the functions have a strong in-class transfer effect. }
  \label{table:detailed_reject_transferable_appendix}
\end{table*}

\subsection{Comparison of different models } On page 6 (~\cref{tab:more_models}), we presented results of using models with different pre-training tasks and inductive biases. Detailed results are presented here in~\cref{tab:more_models_detailed}. These results show that irrespective of the pre-training tasks and inductive biases, neural models perform very similarly. Particularly surprising result is that for models with recurrent inductive bias (XLNet, ESIM-*) which are trained with explicit word order based inductive bias: agreement scores are high even for lexical transformations which destroy examples by changing the word order of the inputs in irregular ways.

\begin{table*}[ht]
  \centering
  \begin{tabular}{lllrrrr}
    \toprule
  \textbf{Class} & \textbf{Pretraining} & \textbf{Model} & \textbf{Lexical} & \textbf{Gradient} & \textbf{PBSMT} & \textbf{Average}\\
    \midrule
    Auto-Encoding            & W-MLM + NSP          & BERT-B~\cite{devlin2019bert} & 80.48 & 57.01 & 55.02        & 67.13               \\
                  &                      & BERT-L   &80.95 & 59.92 & 58.01       & 69.05               \\
                  & W-MLM                & RoBERTa-B ~\cite{liu2019roberta} & 82.35 & 65.63 & 57.02     & 71.72               \\
                  &                      & RoBERTa-L  & 82.90 & 68.17 & 57.42    & 73.52               \\
                  & W-MLM + SOP          & ALBERT-B~\cite{lan2019albert}   & 78.55 & 57.05 & 56.73    & 66.57               \\
    \midrule                      
    Auto-Encoding            & S-MLM + NSP          & SpanBERT-B~\cite{joshi2020spanbert}  & 81.03 & 57.11 & 55.4     & 67.55               \\
                  & S/W-MLM + SS         & BART-B~\cite{DBLP:conf/acl/LewisLGGMLSZ20} & 78.68 & 64.81 & 56.49 &  70.04               \\
                  & RTD                  & ELECTRA-B~\cite{clark2019electra} & 79.36 & 60.95 & 58.11       & 68.81               \\
    \midrule                      
    Auto-Regressive            & P-LM                 & XLNet-B~\cite{yang2019xlnet}   & 81.33 & 62.13 & 57.42        & 70.14               \\
                  & LM                   & ESIM- ELMo~\cite{peters2018deep}   & 76.49 & 78.15 & 63.17    & 75.78               \\    
                  & -                    & ESIM- Glove~\cite{chen2017enhanced}   & 74.19 & 75.31 & 63.41   & 73.49               \\

    \bottomrule
  \end{tabular}
  \caption{Agreement score of different models on MNLI. \textbf{B} refers to the base variant, \textbf{L} refers to the large one. Last four columns present agreement scores for 3 kinds of transformations discussed in this work, with last column being the average. W-MLM refers to word based Masked Language Modeling (MLM), S-MLM refers to span based MLM, NSP refers to Next Sentence Prediction, SOP refers to Sentence Order Prediction, SS refers to sentence shuffling objective, RTD refers to Replaced Token Detection. P-LM refers to permutation language modeling, LM refers to traditional auto-regressive language modeling pre-training objective used in ELMo.}
  \label{tab:more_models_detailed}
\end{table*}

\section{Reproducibility}
In this section, we provide details on our hyperparameter settings along with some comments on reproducibility. 

\begin{table}[]
\centering
\begin{tabular}{l|c}
\toprule
Dataset and Model & Average RunTime \\ \midrule
MNLI + Invalid    & 14 hr           \\
SNLI + Invalid    & 18 hr           \\
MRPC + Invalid    & 13 mins         \\
QQP + Invalid     & 11 hr           \\
SST-2 + Invalid   & 1.5 hr          \\ \bottomrule
\end{tabular}
\caption{Average Training time with an additional invalid class}
\label{tab:runtime}
\end{table}

\paragraph{Models Used} As described in main paper, we performed our experiments by fine-tuning RoBERTa-\textit{base} model \cite{liu2019roberta}. For our implementation, we used Huggingface's \textit{transformers} repository \cite{wolf-etal-2020-transformers}. We also used RoBERTa-\textit{large}, BERT-\textit{base}, BERT-\textit{large} from this repository to compare different models from BERT family.

\paragraph{Hyperparameter Tuning for Calibration Methods}
We used three methods for improving calibration of our model -- \textit{Label Smoothing} (LS), \textit{Focal Loss} (FL), and \textit{Temperature Scaling} (TS). Each of these methods introduce a single new hyperparameter. For Label Smoothing, this parameter ($\lambda_{LS}$) controls the smoothing of hard labels. Focal loss adds a focusing parameter $\gamma_{FL}$ \cite{lin2017focal}. For temperature scaling, we have an additional parameter $T_{TS}$ that controls the amount of scaling to be applied to softmax logits \cite{guo2017calibration}. 

Since all our experiments are on validation datasets, we cannot use those for searching hyperparameters. To find best hyperparameters, we sample 10 \% examples from original training set and use this set for validation. After finding best hyperparameters, the models are trained on original, full training sets.

To find $\lambda_{LS}$ for MNLI, we used grid search over $[0.1, 0.3]$ with 0.05 as step size, our best hyperparameter was $\lambda_{LS} = 0.1$. As LS needs to train a model from scratch for each value of $\lambda_{LS}$, to avoid excessive computation we used this same value across all datasets. For $\gamma_{FL}$, as suggested by ~\citeauthor{lin2017focal}, we use grid search over $[0, 5.0]$ with step size 0.5 and find that $\gamma_{FL} = 2.0$ works best. Again to avoid excessive computation, we fix this value for all datasets. For $T_{TS}$, we maximize the log-likelihood of validation set (described above) using LBFGS optimizer as recommended by ~\citeauthor{guo2017calibration}. Value for $T_{TS}$ was 3.21. We found that while LS, FL improve calibration for all datasets, temperature scaling made calibration worse for SST-2 ($0.05 \to 0.24$).

\paragraph{Computing Infrastructure Used}
Most of our experiments (except on PBSMT) required access to GPU accelerators. We primarily ran our experiments on three machines: Nvidia Tesla V100 (16 GB VRAM) , Tesla P100 (16 GB VRAM), and Nvidia TITAN X (Pascal) (12 GB VRAM). 

We used the Moses SMT system for our PBSMT models~\cite{koehn2007moses,koehn2003statistical}. These required intensive CPU computations and were executed on a machine with an Intel(R) Xeon(R) 2.40GHz CPU  with 28 cores. To speed up computation, we used 48 parallel threads.

\paragraph{Average Run times}
We fine-tune all our models for three epochs with a batch size of 8 (discussed in next section). Average training times are presented in \cref{tab:runtime}.

\paragraph{Fine-tuning Details}

We used the RoBERTa-\textit{base} model for most of our experiments. This model has 12 layers each with hiddem size of 768 and number of attention heads equal to 12. Total number of parameters in this model is 125 million. 

We also experimented with BERT-\textit{base}, BERT-\textit{large}, and RoBERTa-\textit{large} for comparison, which have 110 million, 340 million, and 355 million parameters respectively. For both BERT models, we used the \textit{uncased} variant.

\end{document}